\documentclass{Interspeech}

\usepackage{color, colortbl}

\definecolor{Gray}{gray}{0.95}
\definecolor{Light}{gray}{0.98}
\interspeechcameraready

\title{PAEFF: Precise Alignment and Enhanced Gated Feature Fusion for Face-Voice Association}

\author[affiliation={1}]{Abdul}{Hannan}
\author[affiliation={2}]{Muhammad Arslan}{Manzoor}
\author[affiliation={3}]{Shah}{Nawaz}
\author[affiliation={4}]{Muhammad Irzam}{Liaqat}
\author[affiliation={3,5}]{Markus}{Schedl}
\author[affiliation={2}]{Mubashir}{Noman}

\affiliation{}{University of Trento}{Italy}
\affiliation{}{Mohamed bin Zayed University of Artificial Intelligence}{U.A.E.}
\affiliation{}{Johannes Kepler University}{Austria}
\affiliation{}{IMT Lucca}{Italy}
\affiliation{}{Linz Institute of Technology, AI Lab}{Austria}
\email{}

\usepackage{comment}
\begin{document}

\maketitle

\begin{abstract}
We study the task of learning association between faces and voices, which is gaining interest in the multimodal community lately.
These methods suffer from the deliberate crafting of negative mining procedures as well as the reliance on the distant margin parameter. These issues are addressed by learning a joint embedding space in which orthogonality constraints are applied to the fused embeddings of faces and voices. However, embedding spaces of faces and voices possess different characteristics and require spaces to be aligned before fusing them. To this end, we propose a method that accurately aligns the embedding spaces and fuses them with an enhanced gated fusion thereby improving the performance of face-voice association. Extensive experiments on the VoxCeleb dataset reveals the merits of the proposed approach\footnote[1]{Our code is available at: \textcolor{blue}{https://github.com/hannabdul/paeff}}. 
%
\end{abstract}

\section{Introduction}
\label{sec:intro}
Face-voice association is a widely studied task in cognitive science, which investigates the relationship of the human faces with their voices~\cite{kamachi2003putting,belin2004thinking,rosenblum2006hearing}. 
This task is brought into the computer vision community by Nagrani et al.~\cite{nagrani2018seeing} by formulating a deep learning approach to identify which face pairs belong to the voice segment. 
Consequently, several methods ~\cite{nagrani2018learnable,horiguchi2018face,wen2018disjoint,nawaz2019deep,wen2021seeking,saeed2022fusion,saeed2023single,saeed2024synopsis,shah2023speaker} have been proposed to tackle this problem. These methods rely on unimodal encoders to separately extract the face and voice embeddings followed by learning the discriminative space which minimizes the distance between the embeddings of same identities. 
Typically, contrastive or triplet loss formulations are leveraged to achieve this task~\cite{nagrani2018learnable,horiguchi2018face,nagrani2018seeing,nawaz2021cross,chen2023local}. Although demonstrating favorable performance in learning discriminative space, such loss formulations are limiting in the sense that they require tuning of a margin hyperparameter which is challenging as the distances between samples can differ significantly during training. Furthermore, the training complexities of contrastive and triplet losses are $O(n^2)$ and $O(n^3)$, respectively (where $n$ is the number of samples for a face or voice) resulting in longer training time as the size of the dataset increases dramatically. 
To this end, several methods have explored the utilization of auxiliary identity information as an alternative to contrastive or triplet loss formulations. For example, Nawaz et al.~\cite{nawaz2019deep} introduced an approach to learn the discriminative embedding space by leveraging identity centers, eliminating the need of pairwise or triplet supervision. 
These methods are fairly effective, and require the necessity of being utilized with traditional classification losses such as cross-entropy (CE) loss, resulting in counterintuitive combination of Euclidean and angular spaces respectively. 
To mitigate this, Saeed et al.~\cite{saeed2022fusion} have proposed the fusion and orthogonal projection (FOP) method that introduces a loss by imposing orthogonality constraints on the fused embeddings of faces and voices to learn discriminative embedding space. FOP performs favorably on the face-voice association task, however, fusing the features of two modalities without aligning the corresponding embedding spaces restricts the performance. 


In this work, we hypothesize that the fusion of feature representations that come from different embedding spaces is ineffective unless the two embedding spaces are aligned, and it limits the ability of loss formulation to maximize the distance between the inter-identity representations. 
Therefore, a mechanism is required that can align the embedding spaces of two modalities before fusing them. 
To this end, we propose a dual branch face-voice association framework that enables the model to align the embeddings of faces and voices in hyperbolic space by preserving the distances and complex relationships before fusing them. In addition, we propose a fusion module that can effectively fuse the feature representations resulting in performance improvement of face-voice association tasks. 
In summary, our contributions are as follows.

\begin{itemize}
    \item We propose a dual branch face-voice association framework that precisely aligns the multimodal feature embeddings before merging them.
    \item We introduce an effective feature fusion strategy that utilizes the attention weights to merge face and voice feature embeddings, thereby improving model performance.
    \item We perform extensive experiments on a challenging and benchmark face-voice association dataset (VoxCeleb~\cite{nagrani2017voxceleb}) to demonstrate the effectiveness of the proposed framework.
\end{itemize}

\begin{figure*}[!t]
    \centering
    \includegraphics[width=1.0\linewidth]{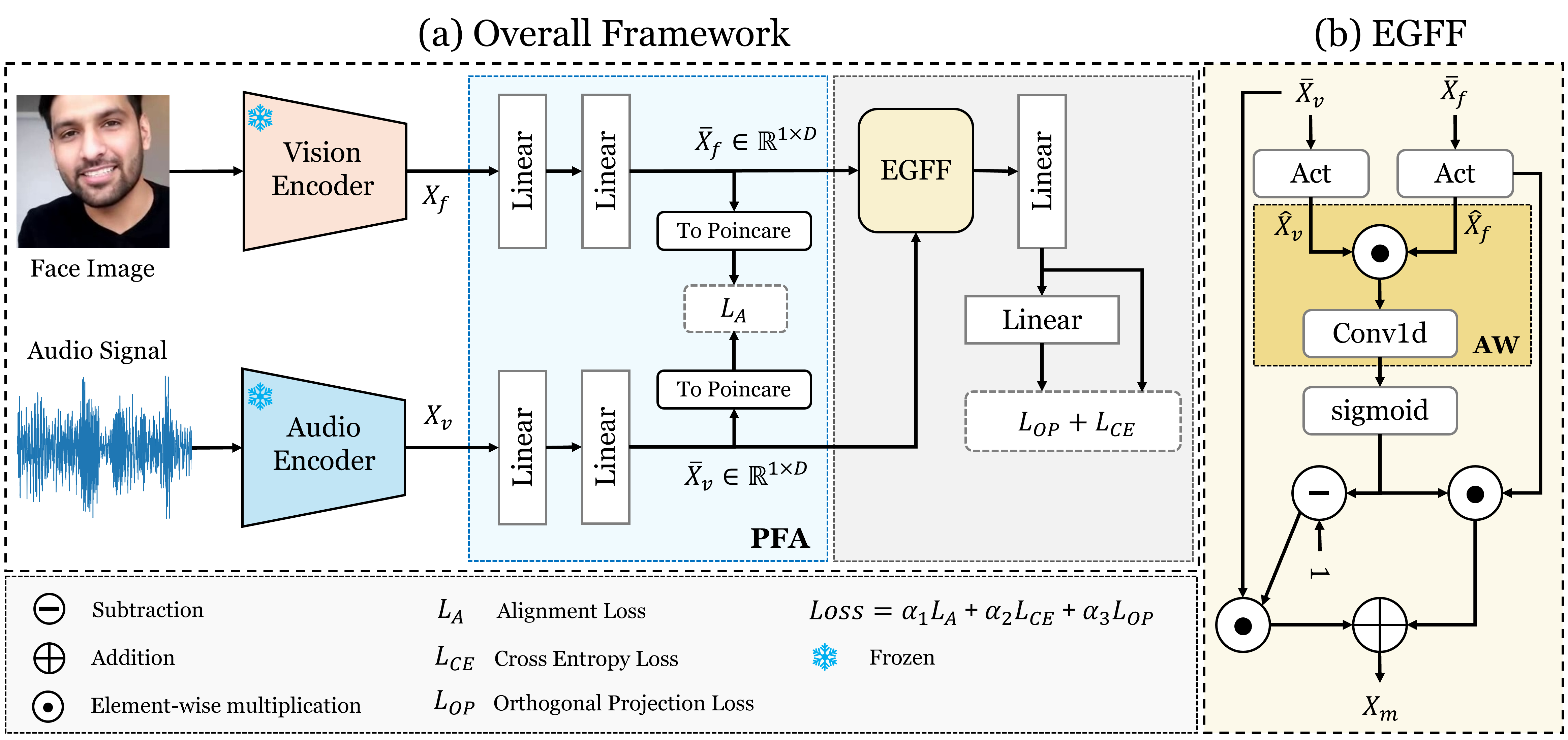}
    \caption{ (a) Overall illustration of the proposed face-voice association approach. Face ($X_f$) and voice ($X_v$) features are extracted by utilizing vision and audio encoders, respectively. Extracted features $X_f$ and $X_v$ are then fed to linear layers to obtain the projected features of dimension $D$. The projected features are transformed to hyperbolic space ($\mathbb H_2$) for accurate alignment of feature representations. Symmetric cross-entropy loss $L_{A}$ is utilized to align the feature representations. Afterwards, aligned features are fused by using enhanced gated feature fusion (EGFF) module. The fused features are fed to the logits layer and optimized by means of orthogonal projection ($L_{OP}$) and cross-entropy ($L_{CE}$) losses. (b) Detailed overview of EGFF is shown on the right side. AW refers to the computation of attention weights for feature fusion. }
    \label{fig:overall_framework}
\end{figure*}


\section{Background}
This section summarizes previous work related to multimodal learning, face-voice association and hyperbolic embedding space.

\noindent \textbf{Multimodal learning and face-voice association.} 
Multimodal learning has been an important research area in recent years due to the fact that information in real-world scenarios comes through multiple and diverse data sources~\cite{baltruvsaitis2018multimodal,zhang2020multimodal,xu2023multimodal,popattia2022guiding,tu2023dctm}.
For instance, human behaviors, emotions, actions, and humor consist of multiple modalities such as speech, spoken language, and faces, thus various human-centered tasks are extensively studied including face-voice association~\cite{horiguchi2018face}.
Nagrani et al.~\cite{nagrani2018seeing,nagrani2018learnable} illustrated that the distance between face and voice embeddings can be minimized by using either contrastive or triplet loss formulation to learn face-voice association. 
Later, Nawaz et al.~\cite{nawaz2019deep} proposed a method to learn a discriminative embedding space by utilizing identity centers, removing the requirement for pairwise or triplet supervision.
More recently, Saeed et al.~\cite{saeed2022fusion} utilized the fused embeddings and employed the orthogonality constraints to minimize the distance between the feature embeddings of same classes. 
Since face and voice feature representations belong to different embedding spaces, therefore, we propose to align these embeddings to obtain discriminative embedding space. 

\noindent \textbf{Hyperbolic Space.} The aim of embedding a space into another requires the preservation of complex relationships of the feature representations. Sala et al.~\cite{sala2018representation} showed that hyperbolic space can better preserve distances and complex relationships than Euclidean space. The authors explained how hyperbolic spaces are related to Poincare disk $\mathbb H_{2}$ that contains all the points in a unit disk in two dimensions.
Later, Khrulkov~\cite{khrulkov2020hyperbolic} showed that hyperblic embeddings are suitable for various computer vision tasks including image classification, image retrieval, and few-shot learning. 
More recently, hyperbolic space is utilized for various multimodal tasks~\cite{ramasinghe2024accept,long2023cross,ibrahimi2024intriguing}.
Building on this insight, we utilized hyperbolic space to effectively model semantic similarities between faces and voices, enhancing the performance of face-voice association task.


\section{Method}
\subsection{Baseline Approach} 
In this work, we adapt a two-branch framework as our baseline method to establish face-voice association~\cite{saeed2022fusion}.
The baseline method utilizes the pre-trained networks to extract features of faces and voices. Afterwards, an attention-based fusion module is used to merge the feature embeddings. These fused embeddings encapsulate the semantics of the identities and orthogonality constraint is applied to maximize the separability between the different identities while minimizing the intra-identity separation. This utilization of orthogonality on fused embeddings provides fair improvement for face-voice association task. 
However, feature representations of faces and voices belong to different embedding spaces. Therefore, it is crucial to align the embedding spaces of faces and voices before merging them. 
To this end, we propose an approach that precisely aligns the feature embeddings in hyperbolic space before fusing them,  thereby improving performance of the model. We describe our approach in detail in the next section.

\subsection{Overall Architecture}
The overall architecture of the proposed framework is illustrated in Figure~\ref{fig:overall_framework}. The proposed framework extracts the features $X_f$ and $X_v$ of the face and voice inputs, respectively, by using the pre-trained face and voice encoders. Following that, we utilize two linear layers to project the feature embeddings to the same dimension. 
As discussed earlier, feature representations of different modality spaces need to be precisely aligned before feature fusion. 
This alignment encourages the model to put emphasis on the features of similar identities during fusion while complementing the orthogonality constraint to maximize the separability between the different identities. To this end, we make face-voice pairs within each batch and minimize the cosine similarity scores between the irrelevant pairs while maximizing the similarity of the relevant face-voice pairs \cite{radford2021learning}. We utilize symmetric cross-entropy loss $\mathcal{L}_{A}$ to learn the feature alignment between the modality spaces. 
Moreover, noticing that the hyperbolic space $\mathbb H_2$ can preserves the distance and complex relationships better as compared to Euclidean space \cite{sala2018representation}, we first project the feature embeddings via Poincaré to hyperbolic space before performing feature alignment. This process encourages the model to effectively align the feature spaces. Later on, we fuse the aligned features by using enhanced gated feature fusion (EGFF). The fused features are then fed to a linear layer to obtain projected feature embeddings. Finally, a classification layer is used to project the features for obtaining the prediction logits. Following the prior work \cite{saeed2022fusion}, we utilize the orthogonal projection loss $\mathcal{L}_{OP}$ and cross entropy loss $\mathcal{L}_{CE}$ for training the model. The total loss $\mathcal{L}$ is the combination of alignment, orthogonal projection, and cross entropy losses given as:

\begin{equation}
\label{eq:loss_equ}
    \mathcal{L} = \alpha_{1} \mathcal{L}_{A} + \alpha_{2} \mathcal{L}_{OP} + \alpha_{3} \mathcal{L}_{CE}
\end{equation}

where $\mathcal{L}_{A}$ is the precise feature alignment loss, $\mathcal{L}_{OP}$ is the orthogonal projection loss, $\mathcal{L}_{CE}$ cross-entropy loss, and ($\alpha_{1}, \alpha_{2}, \alpha_{3}$) are the hyperparameters. 
\begin{table}
\caption{ Cross-modal verification results on \textit{seen-heard} and \textit{unseen-unheard} configurations of the proposed method and existing state-of-the-art methods. Best results are highlighted in bold text whereas the second best are underlined. }
\centering
\resizebox{1.0\linewidth}{!}{
\begin{tabular}{l|cc|cc}
\rowcolor{Gray}
\hline
Method  & EER $\downarrow$ & AUC $\uparrow$ & EER $\downarrow$ & AUC $\uparrow$ \\
\cline{2-5}
\rowcolor{Gray}
& \multicolumn{2}{c|}{Seen-Heard} & \multicolumn{2}{c}{Unseen-Unheard} \\
\hline
\hline
DIMNet~\cite{wen2018disjoint}               & -    & -      & \underline{24.9}          & -          \\
Learnable Pins~\cite{nagrani2018learnable}  & 21.4 & 87.0   & 29.6          & 78.5        \\
MAV-Celeb~\cite{nawaz2021cross}             & -    & -      & 29.0          & 78.9         \\
Single Stream Net.~\cite{nawaz2019deep}     & 17.2 & 91.1          &  29.5 & 78.8  \\
Multi-view~\cite{sari2021multi}             & -    & -      &  28.0         & -               \\
AML~\cite{zheng2021adversarial}             & -    & \underline{92.3 }  & -             &  80.6            \\
DRL~\cite{ning2021disentangled}             & -    & -      & 25.0          &  \textbf{84.6}             \\
FOP~\cite{saeed2022fusion}                  & 19.3 & 89.3   & \underline{24.9} & 83.5      \\
SBNet.~\cite{saeed2023single}               & -    & -    & 25.7 & 82.4      \\

\rowcolor{Light}
\hline
PAEFF (Ours)                                        & \textbf{14.3} & \textbf{93.8} & \textbf{22.9} & \underline{84.4}      \\

\hline
\end{tabular}
}
\label{tab:sota}
\end{table}

\subsection{Enhanced Gated Feature Fusion}
Earlier work~\cite{saeed2022fusion} demonstrated that gated feature fusion \cite{arevalo2017gated} provides better performance compared to the linear fusion strategy for the face-voice association task. We therefore selected the gated fusion strategy for fusing the features of two modalities as shown in Figure~\ref{fig:overall_framework}. Given the aligned feature representations $\Bar{X}_f$ and $\Bar{X}_v$ of the face and voice modalities, we first apply the non-linear activation to the feature representations to obtain features $\hat{X}_f$ and $\hat{X}_v$, respectively. Then, element-wise multiplication is performed between $\hat{X}_f$ and $\hat{X}_v$ followed by convolution layer and \textit{sigmoid} activation to obtain the attention features $X_a$. The element-wise multiplication on the aligned features is pivotal as it highlights the significantly aligned feature representations. Finally, the fused features are obtained by using the equations below:

\begin{equation}
\label{eq:fusion_equ}
    \begin{aligned}
        X_a = \sigma(conv(\hat{X}_f \odot \hat{X}_v)) \\
        X_m = X_a \odot \hat{X}_f + (1-X_a) \odot \hat{X}_v
    \end{aligned}
\end{equation}

where $\sigma$ is the sigmoid operation and $\odot$ represents the element-wise multiplication between the operands.

\section{Experiments}
\noindent \textbf{Implementation Details:} In our experiments, we utilize one Quadro RTX $6000$ GPU to train the proposed model for $50$ epochs using a batch-size of $1024$. 
We set the initial learning rate to $2e^{-5}$. 
During training, we utilize the AdamW optimizer with cosine scheduler. We empirically set the hyperparameters $\alpha_{1}$, $\alpha_{2}$, and $\alpha_{3}$ to $0.3$, $0.35$, and $0.35$ respectively.
Following prior work~\cite{saeed2022fusion}, we employed VGGFace~\cite{parkhi2015deep} and Utterance Level Aggregation~\cite{xie2019utterance} encoders to extract face and voice feature embeddings. 

\noindent \textbf{Dataset:} We perform experiments with VoxCeleb\footnote{\url{https://www.robots.ox.ac.uk/~vgg/data/voxceleb/}}, a benchmark dataset to establish face-voice association.
VoxCeleb is an audio-visual dataset consisting of short clips of human speech, extracted from interview videos uploaded to YouTube~\cite{nagrani2017voxceleb}.
The dataset provides over $100,000$ speaking face-tracks from over $20,000$ videos belonging to $1,251$ speakers.
Following prior work~\cite{nagrani2018learnable,nawaz2019deep,saeed2022fusion,saeed2023single}, we used the same train, validation, and test splits to evaluate on seen-heard and unseen-unheard configurations.
The seen-heard split consists of disjoint videos from the same set of speakers while the unheard-unseen split contains disjoint identities.
Face-voice association is typically established using cross-modal verification and cross-modal matching tasks. 
In cross-modal verification, the goal is to determine whether a given audio segment and a face image correspond to the same identity, while cross-modal matching involves comparing the input modality (probe) to other modality with the varying gallery size $n_c$. 

\noindent \textbf{Evaluation Metrics:} Following prior works~\cite{nagrani2018learnable,saeed2022fusion,saeed2023single}, we utilize the \textit{Equal Error Rate} (EER), \textit{Area Under the Curve} (AUC), and \textit{Matching Accuracy} metrics to evaluate the performance on cross-modal verification and matching tasks.

\begin{figure}
    \centering
    \includegraphics[width=1.0\linewidth]{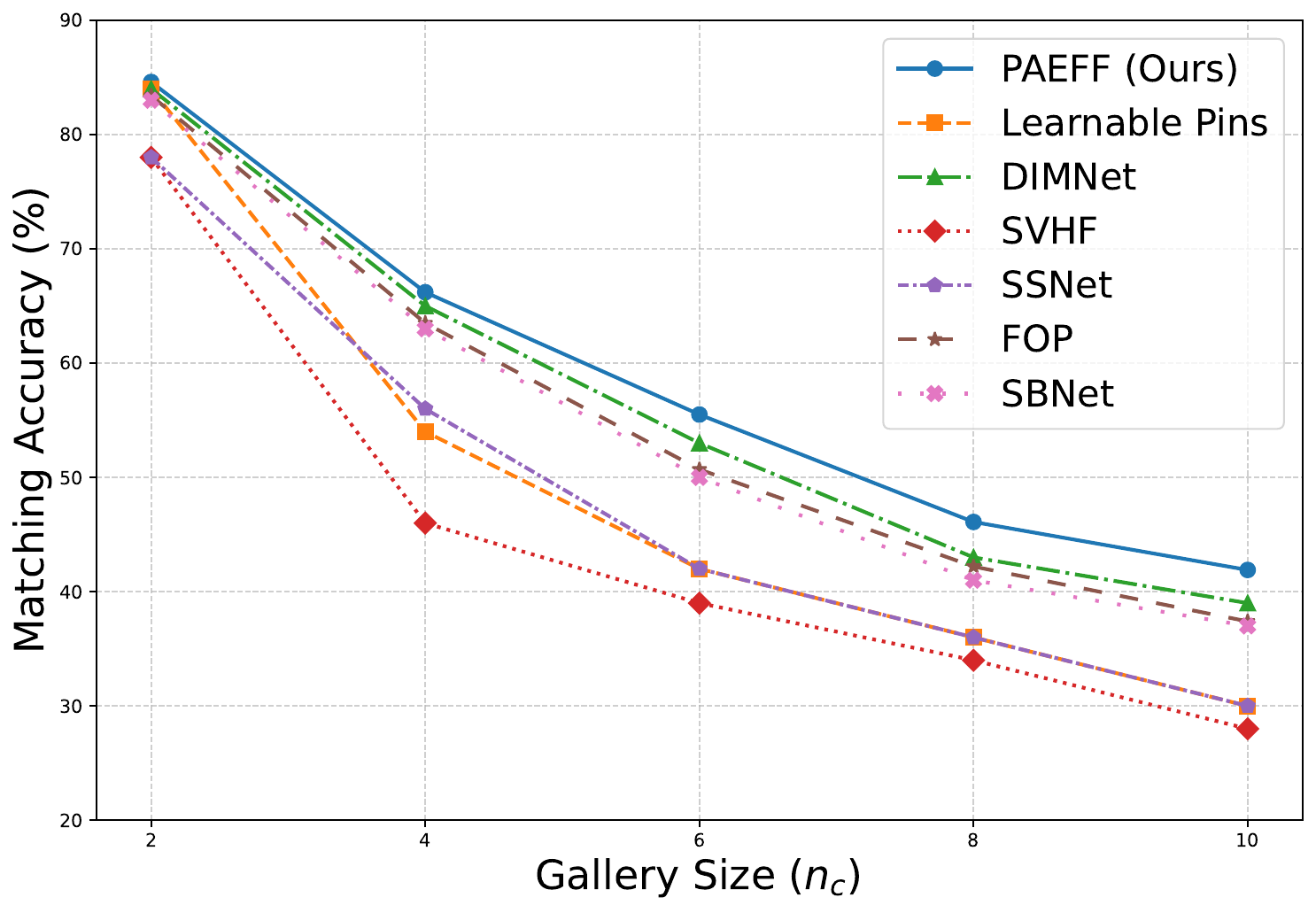}
    \caption{Cross-modal matching results of the proposed model and existing SOTA methods with varying gallery size ($n_c$).}
    \label{fig:sota_matching}
    \vspace{-2.0em}
\end{figure}

\subsection{Face-Voice Association Results}
\noindent \textbf{Cross-modal verification:} We compare the performance of the proposed method against several state-of-the-art (SOTA) face-voice association methods including DIMNet~\cite{wen2018disjoint},  Learnable Pins~\cite{nagrani2018learnable}, MAV-Celeb~\cite{nawaz2021cross}, Single Stream Network~\cite{nawaz2019deep}, Multi-view~\cite{sari2021multi},  Adversarial Metric learning~\cite{zheng2021adversarial} (AML), Disentangled Representation Learning~\cite{ning2021disentangled} (DRL), FOP~\cite{saeed2022fusion}, and Single Branch Network~\cite{saeed2023single} (SBNet).  
Table~\ref{tab:sota} shows the results of cross-modal verification task for both seen-heard and unseen-unheard splits. 
We observe that the proposed method outperforms the other SOTA methods in terms of EER metric by achieving the EER scores of $22.9$\% and $14.3$\%, respectively, for unseen-unheard and seen-heard splits whereas the existing best FOP~\cite{saeed2022fusion} obtains the EER score of $24.9$\% and $19.3$\%, respectively. In terms of AUC metric, our method obtains the favorably better score of $93.8$\% as compared to the AML~\cite{zheng2021adversarial} that achieves the score of $92.3$\% on seen-heard split. However, the proposed method provides on-par performance for unseen-unheard configuration and achieves the AUC score of $84.4$\%.

\noindent \textbf{Cross-modal matching:} In addition to the cross-modal verification task, we compare the performance of the proposed model with SOTA methods for face-voice matching task in Figure~\ref{fig:sota_matching}. It is evident from the results that the proposed model has better matching performance (shown in blue color) as compared to the SOTA method DIMNet \cite{wen2018disjoint} (highlighted in green color) for all gallery sizes ($n_c$ where $c$ in [$2$, $4$, $6$, $8$, $10$]). 
The proposed method achieves an accuracy of $84.6$\% for gallery size $2$ and $41.9$\% for gallery size 10. 
Notably, our method provides significantly better matching performance when the gallery size increases to $8$ or $10$ whereas performance of the other methods degrades considerably for the higher gallery size.

\begin{table}
\caption{ Cross-modal verification results (AUC\%) under varying demographics for \textit{seen-heard} and \textit{unseen-unheard} configurations. Following prior work~\cite{nagrani2018learnable}, rand. configuration is obtained without imposing any demographic restriction. In contrast, G, N, A, GNA are derived by applying their respective demographic constraints. \textdagger Values are taken from prior work~\cite{saeed2022fusion}.
Best results are highlighted in bold text. }
\centering
\resizebox{1.0\linewidth}{!}{%
\begin{tabular}{l|lcccc|lcccc}
\hline
\rowcolor{Gray}
& \multicolumn{10}{|c}{Demographic} \\
\cline{2-11}
\rowcolor{Gray}
Method\textdagger & Rand. & G & N & A & GNA & Rand. & G & N & A & GNA \\
\cline{2-11}
\rowcolor{Gray}
 & \multicolumn{5}{c|}{Seen-Heard}  & \multicolumn{5}{c}{Unseen-Unheard}\\
\hline
CE & 86.6 & 78.0 & 85.0 & 86.3 & 77.3 & 81.7 & 65.9 & 53.6 & 76.0 & 52.8 \\
Center & 88.6  & 79.2   & 87.0  &  88.2  & 78.1  & 77.5  & 62.4  & 51.7  &  72.5  & \textbf{54.2} \\
Git  & 88.9  & 79.7 & 87.4  &  88.6 & 78.5 & 77.9  & 62.6  & 51.8  &  72.8  & \textbf{54.2}   \\
Contrastive  & 84.7  & 69.7     & 83.7     &   84.5    & 69.2   & 79.5  & 61.0     & 53.5     &   74.7    & 51.8  \\
Triplet  & 88.0  & 76.3   & 86.7  &  87.6  & 75.6 & 81.7  & 65.5  & 53.4  & 76.3   & 52.2  \\
FOP      & 89.3  & 76.7  & 87.9 & 88.6 & 76.6  & 83.5 & 68.8  & \textbf{54.9}  & 78.1  & \textbf{54.2}  \\
\rowcolor{Light}
\hline
PAEFF (Ours)  & \textbf{93.8} & \textbf{87.1} & \textbf{92.3} & \textbf{93.4} & \textbf{86.3} & \textbf{84.4}  & \textbf{71.5}  & 53.0 & \textbf{78.8} & 54.0  \\

\hline
\end{tabular}}
\label{tab:results-demographic} 
\end{table}

\noindent \textbf{Comparison on varying demographic combinations: } 
Table~\ref{tab:results-demographic} presents the effect of Gender (G), Nationality (N), Age (A), and their combination (GNA) separately on face-voice association problem. 
We notice that the proposed method consistently outperforms the other approaches in terms of AUC score for the seen-heard split of all demographics achieving a considerable gain of 7.8\% for GNA combination. For unseen-unheard configuration, the proposed approach performs favorably for the random, G, and A combinations by obtaining AUC scores of 84.4\%, 71.5\%, and 78.8\%, respectively. However, we observe that the FOP, Center, and Git loss approaches provide better performance on N and GNA combinations of unseen-unheard split. 

\subsection{Ablation Study}
We demonstrate the effectiveness of each of the proposed components in Table~\ref{tab:ablation}. 
For the unseen-unheard configuration, the baseline approach achieves EER and AUC scores of $27.6$\% and $79.2$\%, respectively. 
Replacing the gated fusion mechanism in the baseline method with the proposed EGFF module provides an AUC gain of $1.5$\% and reduces the EER score by $1.6$\%. Similarly, aligning the embedding spaces of two modalities before feature fusion leads to significant improvement in the AUC and EER metrics achieving scores of $83.6$\% and $23.9$\%, respectively. Finally, projecting the feature representations to the hyperbolic space and aligning the embedding spaces further improves the face-voice association performance and achieves the state-of-the-art results of $22.9$\% in terms of EER metrics, underscoring the merits of the proposed contributions. 

Table~\ref{tab:ablation} also reports the performance comparison of the proposed components on seen-heard split. We observe the similar performance improvement when the proposed components are integrated in the baseline approach. In particular, we notice that there is a significant reduction of $8.9$\% in EER score and considerable gain of $8.0$\% in AUC when the embedding spaces of the two modalities are aligned before feature fusion. However, the reduction in EER score is not considerable when feature representations are aligned in the hyperbolic space in case of seen-heard configuration.


\noindent \textbf{Computation of Attention Weights: } The attention weights utilized to highlight the relevant semantic features in EGFF (Figure~\ref{fig:overall_framework}) are crucial. Therefore, we investigate how to combine face and voice features for obtaining rich semantic information. Table~\ref{tab:ablation_fusion} presents the results of obtaining attention weights by utilizing addition, concatenation, and multiplication operations for feature fusion. We observe that the AUC score of the proposed approach is not affected by the type of operation used for computing attention weights. 
However, the multiplication operation is better for computing attention weights when face and voice features are fused using gated feature fusion mechanism, improving EER score.

\begin{table}[t]
\caption{Ablation study demonstrating the effectiveness of proposed components on both seen-heard and unseen-unheard configurations. 
FA represents the feature alignment in Euclidean space before fusion. 
Best results are highlighted in bold text.}
\centering
\resizebox{01.0\linewidth}{!}{%
\begin{tabular}{l|cc|cc}
\rowcolor{Gray}
\hline
Method   & EER $\downarrow$ & AUC $\uparrow$ & EER $\downarrow$ & AUC $\uparrow$ \\
\cline{2-5}
\rowcolor{Gray}
 & \multicolumn{2}{c|}{Seen-Heard}  & \multicolumn{2}{c}{Unseen-Unheard}\\
\hline
Baseline & 26.5 & 81.0 & 27.6 & 79.2 \\
Baseline + EGFF & 23.8 & 84.1 & 26.0 & 80.7  \\
Baseline + EGFF + FA & 14.9 & 92.1 & 23.9 & 83.6  \\
\rowcolor{Light}
\hline
PAEFF (Ours) & \textbf{14.3} & \textbf{93.8} & \textbf{22.9} & \textbf{84.4} \\

\hline
\end{tabular}
}
\label{tab:ablation}
\end{table}

\begin{table}[t]
\caption{Ablation of computing the attention weights for fusion of face and voice features on unseen-unheard configuration. 
Best results are highlighted in bold text.}
\centering
\setlength{\tabcolsep}{20pt}
\begin{tabular}{l|cc}
\rowcolor{Gray}
\hline
Method   & EER $\downarrow$ & AUC $\uparrow$ \\
\hline
Addition & 24.4 & \textbf{84.4}  \\
Concatenation & 24.0 & 84.3 \\
\rowcolor{Light}

Multiplication & \textbf{22.9} & \textbf{84.4} \\
\hline
\end{tabular}
\label{tab:ablation_fusion}
\end{table}

\section{Conclusion}
In this work, we introduced an effective approach that accurately aligns the face-voice embeddings before fusion. We demonstrated that the precise alignment of features is a crucial step for obtaining better performance in the face-voice association task. Moreover, we showed that effectively fusing the face-voice feature representations also improves the results of the face-voice association task. 
Extensive experiments conducted on the VoxCeleb benchmark reveal the merits of the proposed approach. 
Our potential future direction is to explore the adaptation of the proposed approach for multilingual face-voice association and speaker diarization.



\bibliographystyle{IEEEtran}
\bibliography{mybib}

\begin{thebibliography}{10}
\providecommand{\url}[1]{#1}
\csname url@samestyle\endcsname
\providecommand{\newblock}{\relax}
\providecommand{\bibinfo}[2]{#2}
\providecommand{\BIBentrySTDinterwordspacing}{\spaceskip=0pt\relax}
\providecommand{\BIBentryALTinterwordstretchfactor}{4}
\providecommand{\BIBentryALTinterwordspacing}{\spaceskip=\fontdimen2\font plus
\BIBentryALTinterwordstretchfactor\fontdimen3\font minus \fontdimen4\font\relax}
\providecommand{\BIBforeignlanguage}[2]{{%
\expandafter\ifx\csname l@#1\endcsname\relax
\typeout{** WARNING: IEEEtran.bst: No hyphenation pattern has been}%
\typeout{** loaded for the language `#1'. Using the pattern for}%
\typeout{** the default language instead.}%
\else
\language=\csname l@#1\endcsname
\fi
#2}}
\providecommand{\BIBdecl}{\relax}
\BIBdecl

\bibitem{kamachi2003putting}
M.~Kamachi, H.~Hill, K.~Lander, and E.~Vatikiotis-Bateson, ``Putting the face to the voice': Matching identity across modality,'' \emph{Current Biology}, vol.~13, no.~19, pp. 1709--1714, 2003.

\bibitem{belin2004thinking}
P.~Belin, S.~Fecteau, and C.~Bedard, ``Thinking the voice: neural correlates of voice perception,'' \emph{Trends in cognitive sciences}, vol.~8, no.~3, pp. 129--135, 2004.

\bibitem{rosenblum2006hearing}
L.~D. Rosenblum, N.~M. Smith, S.~M. Nichols, S.~Hale, and J.~Lee, ``Hearing a face: Cross-modal speaker matching using isolated visible speech,'' \emph{Perception \& psychophysics}, vol.~68, pp. 84--93, 2006.

\bibitem{nagrani2018seeing}
A.~Nagrani, S.~Albanie, and A.~Zisserman, ``Seeing voices and hearing faces: Cross-modal biometric matching,'' in \emph{Proceedings of the IEEE conference on computer vision and pattern recognition}, 2018, pp. 8427--8436.

\bibitem{nagrani2018learnable}
A.~Nagrani, S.~Albanie, and A.~Zisserman, ``Learnable pins: Cross-modal embeddings for person identity,'' in \emph{Proceedings of the European conference on computer vision (ECCV)}, 2018, pp. 71--88.

\bibitem{horiguchi2018face}
S.~Horiguchi, N.~Kanda, and K.~Nagamatsu, ``Face-voice matching using cross-modal embeddings,'' in \emph{Proceedings of the 26th ACM international conference on Multimedia}, 2018, pp. 1011--1019.

\bibitem{wen2018disjoint}
Y.~Wen, M.~A. Ismail, W.~Liu, B.~Raj, and R.~Singh, ``Disjoint mapping network for cross-modal matching of voices and faces,'' in \emph{7th International Conference on Learning Representations, {ICLR} 2019, USA, May 6-9, 2019}, 2019.

\bibitem{nawaz2019deep}
S.~Nawaz, M.~K. Janjua, I.~Gallo, A.~Mahmood, and A.~Calefati, ``Deep latent space learning for cross-modal mapping of audio and visual signals,'' in \emph{2019 Digital Image Computing: Techniques and Applications (DICTA)}.\hskip 1em plus 0.5em minus 0.4em\relax IEEE, 2019, pp. 1--7.

\bibitem{wen2021seeking}
P.~Wen, Q.~Xu, Y.~Jiang, Z.~Yang, Y.~He, and Q.~Huang, ``Seeking the shape of sound: An adaptive framework for learning voice-face association,'' in \emph{Proceedings of the IEEE/CVF conference on computer vision and pattern recognition}, 2021, pp. 16\,347--16\,356.

\bibitem{saeed2022fusion}
M.~S. Saeed, M.~H. Khan, S.~Nawaz, M.~H. Yousaf, and A.~Del~Bue, ``Fusion and orthogonal projection for improved face-voice association,'' in \emph{ICASSP 2022-2022 IEEE International Conference on Acoustics, Speech and Signal Processing (ICASSP)}.\hskip 1em plus 0.5em minus 0.4em\relax IEEE, 2022, pp. 7057--7061.

\bibitem{saeed2023single}
M.~S. Saeed, S.~Nawaz, M.~H. Khan, M.~Z. Zaheer, K.~Nandakumar, M.~H. Yousaf, and A.~Mahmood, ``Single-branch network for multimodal training,'' in \emph{ICASSP 2023-2023 IEEE International Conference on Acoustics, Speech and Signal Processing (ICASSP)}.\hskip 1em plus 0.5em minus 0.4em\relax IEEE, 2023, pp. 1--5.

\bibitem{saeed2024synopsis}
M.~S. Saeed, S.~Nawaz, M.~Moscati, R.~K. Das, M.~S. Tahir, M.~Z. Zaheer, M.~I. Liaqat, M.~H. Khan, K.~Nandakumar, M.~H. Yousaf \emph{et~al.}, ``A synopsis of fame 2024 challenge: Associating faces with voices in multilingual environments,'' in \emph{Proceedings of the 32nd ACM International Conference on Multimedia}, 2024, pp. 11\,333--11\,334.

\bibitem{shah2023speaker}
S.~H. Shah, M.~S. Saeed, S.~Nawaz, and M.~H. Yousaf, ``Speaker recognition in realistic scenario using multimodal data,'' in \emph{2023 3rd International Conference on Artificial Intelligence (ICAI)}.\hskip 1em plus 0.5em minus 0.4em\relax IEEE, 2023, pp. 209--213.

\bibitem{nawaz2021cross}
S.~Nawaz, M.~S. Saeed, P.~Morerio, A.~Mahmood, I.~Gallo, M.~H. Yousaf, and A.~Del~Bue, ``Cross-modal speaker verification and recognition: A multilingual perspective,'' in \emph{Proceedings of the IEEE/CVF conference on computer vision and pattern recognition}, 2021, pp. 1682--1691.

\bibitem{chen2023local}
G.~Chen, D.~Zhang, T.~Liu, and X.~Du, ``Local-global contrast for learning voice-face representations,'' in \emph{2023 IEEE International Conference on Image Processing (ICIP)}.\hskip 1em plus 0.5em minus 0.4em\relax IEEE, 2023, pp. 51--55.

\bibitem{nagrani2017voxceleb}
A.~Nagrani, J.~S. Chung, and A.~Zisserman, ``Voxceleb: a large-scale speaker identification dataset,'' \emph{arXiv preprint arXiv:1706.08612}, 2017.

\bibitem{baltruvsaitis2018multimodal}
T.~Baltru{\v{s}}aitis, C.~Ahuja, and L.-P. Morency, ``Multimodal machine learning: A survey and taxonomy,'' \emph{IEEE transactions on pattern analysis and machine intelligence}, vol.~41, no.~2, pp. 423--443, 2018.

\bibitem{zhang2020multimodal}
C.~Zhang, Z.~Yang, X.~He, and L.~Deng, ``Multimodal intelligence: Representation learning, information fusion, and applications,'' \emph{IEEE Journal of Selected Topics in Signal Processing}, vol.~14, no.~3, pp. 478--493, 2020.

\bibitem{xu2023multimodal}
P.~Xu, X.~Zhu, and D.~A. Clifton, ``Multimodal learning with transformers: A survey,'' \emph{IEEE Transactions on Pattern Analysis and Machine Intelligence}, vol.~45, no.~10, pp. 12\,113--12\,132, 2023.

\bibitem{popattia2022guiding}
M.~Popattia, M.~Rafi, R.~Qureshi, and S.~Nawaz, ``Guiding attention using partial-order relationships for image captioning,'' in \emph{Proceedings of the IEEE/CVF conference on computer vision and pattern recognition}, 2022, pp. 4671--4680.

\bibitem{tu2023dctm}
V.~N. Tu, V.~T. Huynh, H.-J. Yang, S.-H. Kim, S.~Nawaz, K.~Nandakumar, and M.~Z. Zaheer, ``Dctm: Dilated convolutional transformer model for multimodal engagement estimation in conversation,'' in \emph{Proceedings of the 31st ACM International Conference on Multimedia}, 2023, pp. 9521--9525.

\bibitem{sala2018representation}
F.~Sala, C.~De~Sa, A.~Gu, and C.~R{\'e}, ``Representation tradeoffs for hyperbolic embeddings,'' in \emph{International conference on machine learning}.\hskip 1em plus 0.5em minus 0.4em\relax PMLR, 2018, pp. 4460--4469.

\bibitem{khrulkov2020hyperbolic}
V.~Khrulkov, L.~Mirvakhabova, E.~Ustinova, I.~Oseledets, and V.~Lempitsky, ``Hyperbolic image embeddings,'' in \emph{Proceedings of the IEEE/CVF conference on computer vision and pattern recognition}, 2020, pp. 6418--6428.

\bibitem{ramasinghe2024accept}
S.~Ramasinghe, V.~Shevchenko, G.~Avraham, and A.~Thalaiyasingam, ``Accept the modality gap: An exploration in the hyperbolic space,'' in \emph{Proceedings of the IEEE/CVF Conference on Computer Vision and Pattern Recognition}, 2024, pp. 27\,263--27\,272.

\bibitem{long2023cross}
T.~Long and N.~van Noord, ``Cross-modal scalable hyperbolic hierarchical clustering,'' in \emph{Proceedings of the IEEE/CVF international conference on computer vision}, 2023, pp. 16\,655--16\,664.

\bibitem{ibrahimi2024intriguing}
S.~Ibrahimi, M.~G. Atigh, N.~Van~Noord, P.~Mettes, and M.~Worring, ``Intriguing properties of hyperbolic embeddings in vision-language models,'' \emph{Transactions on Machine Learning Research}, 2024.

\bibitem{radford2021learning}
A.~Radford, J.~W. Kim, C.~Hallacy, A.~Ramesh, G.~Goh, S.~Agarwal, G.~Sastry, A.~Askell, P.~Mishkin, J.~Clark \emph{et~al.}, ``Learning transferable visual models from natural language supervision,'' in \emph{International conference on machine learning}.\hskip 1em plus 0.5em minus 0.4em\relax PMLR, 2021, pp. 8748--8763.

\bibitem{sari2021multi}
L.~Sar{\i}, K.~Singh, J.~Zhou, L.~Torresani, N.~Singhal, and Y.~Saraf, ``A multi-view approach to audio-visual speaker verification,'' in \emph{ICASSP 2021-2021 IEEE International Conference on Acoustics, Speech and Signal Processing (ICASSP)}.\hskip 1em plus 0.5em minus 0.4em\relax IEEE, 2021, pp. 6194--6198.

\bibitem{zheng2021adversarial}
A.~Zheng, M.~Hu, B.~Jiang, Y.~Huang, Y.~Yan, and B.~Luo, ``Adversarial-metric learning for audio-visual cross-modal matching,'' \emph{IEEE Transactions on Multimedia}, vol.~24, pp. 338--351, 2021.

\bibitem{ning2021disentangled}
H.~Ning, X.~Zheng, X.~Lu, and Y.~Yuan, ``Disentangled representation learning for cross-modal biometric matching,'' \emph{IEEE Transactions on Multimedia}, vol.~24, pp. 1763--1774, 2021.

\bibitem{arevalo2017gated}
J.~Arevalo, T.~Solorio, M.~Montes-y G{\'o}mez, and F.~A. Gonz{\'a}lez, ``Gated multimodal units for information fusion,'' \emph{arXiv preprint arXiv:1702.01992}, 2017.

\bibitem{parkhi2015deep}
O.~Parkhi, A.~Vedaldi, and A.~Zisserman, ``Deep face recognition,'' in \emph{BMVC 2015-Proceedings of the British Machine Vision Conference 2015}.\hskip 1em plus 0.5em minus 0.4em\relax British Machine Vision Association, 2015.

\bibitem{xie2019utterance}
W.~Xie, A.~Nagrani, J.~S. Chung, and A.~Zisserman, ``Utterance-level aggregation for speaker recognition in the wild,'' in \emph{ICASSP 2019-2019 IEEE International Conference on Acoustics, Speech and Signal Processing (ICASSP)}.\hskip 1em plus 0.5em minus 0.4em\relax IEEE, 2019, pp. 5791--5795.

\end{thebibliography}

\end{document}